\documentclass[runningheads]{llncs}
\usepackage[T1]{fontenc}
\usepackage{graphicx}
\usepackage{times}
\usepackage{soul}
\usepackage{url}
\usepackage[hidelinks]{hyperref}
\usepackage[utf8]{inputenc}
\usepackage[small]{caption}
\usepackage{graphicx}
\usepackage{amsmath}
\usepackage{booktabs}
\usepackage[switch]{lineno}
\usepackage{placeins}

\usepackage{subcaption}
\usepackage{amssymb}
\usepackage{mathtools}
\usepackage{bm}
\usepackage[capitalize,noabbrev]{cleveref}
\usepackage{color}

\begin{document}
\raggedbottom
\setlength{\floatsep}{8mm plus 2pt minus 2pt}
\makeatletter
\setlength{\@fptop}{0pt}
\makeatother
%


\newcommand{\fixme}[2][]{#2}
\renewcommand{\fixme}[2][]{\textcolor{red}{#2}}
\renewcommand{\fixme}[2][]{\textcolor{red}{#2\footnote{#1}}}

\title{Fast Surrogate Modeling of Excitable and Oscillatory FitzHugh-Nagumo Dynamics with Parametric Neural Operators}

\titlerunning{FitzHugh-Nagumo Dynamics with Parametric Neural Operators}

\author{Andrew Franck\inst{1}\orcidID{0009-0009-6729-4262}
\and Justin Li\inst{1}\orcidID{0000-0002-9660-6570}
}

\authorrunning{A. Franck and J. Li}

\institute{Occidental College, Los Angeles, CA 90041, USA\\
\email{franck@oxy.edu \; justinnhli@oxy.edu}
}

\maketitle

\begin{abstract}
The FitzHugh-Nagumo (FHN) system serves as a simplified model of neuronal voltage dynamics, capturing the activator-inhibitor structure behind both isolated action potentials and the rhythmic spiking seen across the brain. Exploring its 5D physiological parameter space is important for neuromodulation and mapping voltage recordings back to biophysics, yet classical finite-difference solvers make rapid parameter sweeps expensive. We train parameter-conditioned Fourier Neural Operators (FNOs) as fast, differentiable surrogates for the FHN voltage and recovery fields on a one-dimensional spatial domain, conditioning each Fourier layer on the parameter vector $\bm{\lambda} = (D_u, D_v, a, b, \tau)$ via feature-wise linear modulation (FiLM). We apply a single bifurcation analysis that delimits the two distinct regimes the model spans, oscillatory (tonic firing) and excitable (action-potential propagation), and we train one operator in each. In the oscillatory regime the surrogate attains sub-$0.1\%$ relative $L^2$ error on both fields, runs nearly three orders of magnitude faster than the finite-difference baseline, generalizes uniformly across the parameter space, and extrapolates to low single-digit percentage errors outside of the training bounds. In the excitable regime the same operator accurately reproduces the firing threshold and the $c \propto \sqrt{D_u}$ conduction-velocity law and replicates full traveling pulses, fully capturing the excitable bifurcation structure rather than just smoothly interpolating fields.

\end{abstract}

\section{Introduction}

Much of the brain's signaling is shaped by two coupled processes inside each neuron: a fast change in membrane voltage that produces an action potential, and a slower recovery current that resets the cell~\cite{Izhikevich2007,Ermentrout2010}. The FitzHugh-Nagumo (FHN) reaction-diffusion system~\cite{FitzHugh1961,Nagumo1962} reduces these dynamics into two coupled equations for membrane voltage and recovery, and remains a widely used reduced model of neural excitation by capturing both single action-potential propagation and the rhythmic spiking that dominates many human-brain signals. Connecting noisy voltage recordings back to the parameters of such a model is important for neural decoding, inverse modeling, and closed-loop neuromodulation.

These are all tasks that demand many fast, differentiable forward solves, which traditional finite-difference or spectral methods~\cite{Ermentrout2010,2020PhRvR...2b3068B} struggle to provide, as they require fine discretization and become prohibitive over large parameter sweeps. For parametric PDEs, the cost of classical solvers grows quickly with the parameter space, since each new parameter combination requires a full re-simulation. This makes tasks such as inverse parameter inference (fitting $\bm{\lambda}$ to a voltage recording) and stimulus optimization for closed-loop control challenging, and motivates fast learned surrogates. This paper presents a step in this direction by training such surrogates using Neural Operators (NOs)~\cite{kovachki2021neural}. We restrict our attention to a single spatial dimension, leaving the extension of the surrogate to 2D/3D domains to future work.

\subsection*{Contributions:}
\begin{itemize}
    \item We train a parameter-conditioned FNO for the FHN system, using feature-wise linear modulation to add $\bm{\lambda} = (D_u, D_v, a, b, \tau)$ into every Fourier layer. A single network covers the full 5D parameter space, whereas prior reaction-diffusion work~\cite{hao2024fourier} fixed the physical parameters.
    \item A bifurcation analysis of the ODE shows that the sampled parameter ranges sit clearly in the oscillatory and excitable regimes of firing neurons.
    \item We characterize parameter-wise error sensitivity across the 5D space, and demonstrate relatively uniform generalization, with $|\rho| < 0.36$ for all parameter-error correlations.
    \item We quantify extrapolation behavior outside the training ranges and identify diffusion magnitude as the primary failure mode, while the surrogate extrapolates accurately along the reaction and time-scale axes.
    \item Within the training parameter range, the surrogate holds sub-$0.1\%$ relative $L^2$ error on both fields while running nearly three orders of magnitude faster than the finite-difference baseline. This positions the surrogate as a fast, differentiable forward model for downstream neuroscience workloads such as inverse parameter inference and closed-loop stimulation, which we motivate and set up but leave to future work.
    \item We show the same FiLM-conditioned operator extends to the excitable (action-potential) regime, where it reproduces the all-or-none firing threshold and the $c \propto \sqrt{D_u}$ conduction-velocity scaling and replicates full traveling action potentials over $100$-step rollouts.
\end{itemize}

\section{Related Work and Mathematical Background}

\subsection{The FitzHugh-Nagumo System}

The FitzHugh-Nagumo (FHN) system models neuronal voltage and recovery dynamics through two coupled reaction-diffusion equations~\cite{FitzHugh1961,Nagumo1962}:

\begin{align}
\frac{\partial u}{\partial t} &= D_u \nabla^2 u + u - \frac{u^3}{3} - v, \label{eq:fhn_u} \\
\frac{\partial v}{\partial t} &= D_v \nabla^2 v + \frac{1}{\tau}(u + a - b v), \label{eq:fhn_v}
\end{align}

\noindent where $u(\bm{x},t)$ is the membrane voltage, $v(\bm{x},t)$ is the recovery variable that gates the slow current, $\bm{x} \in \Omega \subseteq \mathbb{R}^1$ is the spatial coordinate, and $t \in [0,T]$ is time. The parameter vector $\bm{\lambda} = (D_u, D_v, a, b, \tau) \in \mathbb{R}^5$ characterizes the system, where $D_u, D_v > 0$ are diffusion coefficients and $(a, b, \tau)$ control the reaction kinetics. Different combinations of $\bm{\lambda}$ place the system in qualitatively distinct dynamical regimes---principally the \emph{oscillatory} (tonic-firing) and \emph{excitable} (single action-potential) regimes this paper models. Before generating data, we verify via the bifurcation analysis of \Cref{sec:bifurcation} that each sampled parameter box lies cleanly inside its intended regime.

The FitzHugh-Nagumo model has traditionally been studied with numerical methods such as finite difference methods, finite element methods, and spectral methods \cite{Quarteroni2000,LeVeque2007}. These approaches discretize the spatial domain into grids or elements and then iteratively solve the resulting equations. These methods are accurate but require fine spatial and temporal discretization, making large parameter sweeps computationally expensive.

\subsection{Learning Parametric Solution Operators with FNOs}

One solution to this problem is to use machine-learned surrogate models instead. In particular, Neural Operators (NOs)~\cite{kovachki2021neural} learn mappings between infinite-dimensional function spaces, capturing the solution operator itself rather than a discretization. Fourier Neural Operators~\cite{li2021fourierneuraloperatorparametric} extend the architecture by performing their convolutions in the frequency domain, making them are particularly effective for PDEs such as the FHN system.

Although surrogate models have been widely used, the use of FNOs, to our knowledge, has not yet been applied to FHN systems or similar biological neural models. Existing FNO surrogates instead target problems in the physical sciences: model-parallel FNOs for billion-variable subsurface CO$_2$ flow~\cite{grady2023modelparallel}, surrogate forward solvers for nonlinear electrical resistivity tomography that make Bayesian inversion tractable~\cite{ghadjari2026fno}, and adjoint-style FNO solvers for wavefront shaping in tunable metasurfaces~\cite{kang2024adjoint}; the latter two, like our own motivation, exploit the differentiable surrogate to drive an inverse or optimization loop. The closest prior work, Hao and Song~\cite{hao2024fourier}, applies FNOs to the Surface Quasi-Geostrophic and Gray-Scott systems, but (i) holds physical parameters fixed, (ii) targets smooth field evolution, and (iii) evaluates generalization only across initial conditions. Our system adds stiff dynamics, sharper activations, and a five-dimensional parameter space that controls distinct ODE bifurcations.

We consider the challenge of learning nonlinear parametric operators $\mathcal{G}_{\bm{\lambda}}: \mathcal{U} \times \mathbb{R}^5 \rightarrow \mathcal{V}$ that map initial states to future states of the FHN system across the parameter space. These operators define mappings between infinite-dimensional function spaces $\mathcal{U}$ and $\mathcal{V}$ for each parameter configuration $\bm{\lambda} \in \mathbb{R}^5$. The input functions $\bm{u}_0 = (u_0, v_0): \Omega \rightarrow \mathbb{R}^2$ represent initial conditions, which are transformed by the operator into solution fields $\bm{u}_{\Delta t} = \mathcal{G}_{\bm{\lambda}}[\bm{u}_0]: \Omega \rightarrow \mathbb{R}^2$ at time $t = \Delta t$. We specifically learn a single-step operator that maps the state at time $t$ to the state at time $t + \Delta t$:
\begin{equation*}
    \bm{u}_{t+\Delta t}(\bm{x},t) = \mathcal{G}_{\bm{\lambda}}^{\Delta t}[\bm{u}_t](\bm{x}), \quad \bm{x} \in \Omega.
    \label{eq:single_step}
\end{equation*}
\noindent For longer time predictions, the single-step operator is applied iteratively:
\begin{equation*}
    \bm{u}_{n\Delta t} = \underbrace{\mathcal{G}_{\bm{\lambda}}^{\Delta t} \circ \mathcal{G}_{\bm{\lambda}}^{\Delta t} \circ \cdots \circ \mathcal{G}_{\bm{\lambda}}^{\Delta t}}_{n \text{ times}}[\bm{u}_0].
    \label{eq:autoregressive}
\end{equation*}

\subsection{Fourier Neural Operator Architecture}
\label{sec:fno_architecture}

The Fourier Neural Operator~\cite{li2021fourierneuraloperatorparametric} approximates the solution operator $\mathcal{G}_{\bm{\lambda}}$ through a neural network $\mathcal{G}_{\bm{\theta}, \bm{\lambda}}$, where $\bm{\theta}$ denotes the learnable weights. The architecture consists of three main components:

\begin{enumerate}
    \item Lifting: An initial projection $\mathcal{P}: \mathbb{R}^{d_{\text{in}}} \rightarrow \mathbb{R}^w$ maps the input channels (here $d_{\text{in}} = 2$ for $(u,v)$) to a higher-dimensional representation of width $w$:
        \begin{equation*}
            \bm{z}_0(\bm{x}) = \mathcal{P}(\bm{u}_0(\bm{x})).
        \end{equation*}

    \item Fourier Layers: The core of the FNO consists of $L$ Fourier layers. Each layer $\ell \in \{1, \ldots, L\}$ applies a convolution and then a nonlinearity:
        \begin{equation*}
            \bm{z}_{\ell}(\bm{x}) = \sigma\left(\mathcal{K}_{\ell}[\bm{z}_{\ell-1}](\bm{x}) + \mathcal{W}_{\ell}(\bm{z}_{\ell-1}(\bm{x}))\right)
        \end{equation*}
        where $\sigma$ is a nonlinear activation function, $\mathcal{W}_{\ell}$ is a linear transformation (implemented as $1\times1$ convolution), and $\mathcal{K}_{\ell}$ is the convolution operator. The convolution $\mathcal{K}_{\ell}$ operates by transforming to the frequency domain, multiplying with learnable weights, and transforming back:
        \begin{equation*}
            \mathcal{K}_{\ell}[\bm{z}](\bm{x}) = \mathcal{F}^{-1}\left(\bm{R}_{\ell} \cdot (\mathcal{F}\bm{z})\right)(\bm{x})
        \end{equation*}
        where $\mathcal{F}$ and $\mathcal{F}^{-1}$ are the Fourier transform and its inverse, while $\bm{R}_{\ell} \in \mathbb{C}^{w \times w \times k_{\max}}$ are learnable weights~\cite{li2021fourierneuraloperatorparametric}.

    \item Projection: A projection $\mathcal{Q}: \mathbb{R}^w \rightarrow \mathbb{R}^{d_{\text{out}}}$ maps from the hidden dimension back to the output space ($d_{\text{out}} = 2$ for $(u,v)$):
        \begin{equation*}
            \bm{u}_{\Delta t}(\bm{x}) = \mathcal{Q}(\bm{z}_L(\bm{x})).
        \end{equation*}
\end{enumerate}

\subsection{Training Objective}
\label{sec:training}

Given a dataset $\mathcal{D} = \{(\bm{u}_0^{(i)}, \bm{u}_{\Delta t}^{(i)}, \bm{\lambda}^{(i)})\}_{i=1}^N$ of initial conditions, solutions, and parameters, we minimize the mean squared error:
\begin{equation*}
    \mathcal{L}(\bm{\theta}) = \frac{1}{N} \sum_{i=1}^N \|\mathcal{G}_{\bm{\theta}, \bm{\lambda}^{(i)}}[\bm{u}_0^{(i)}] - \bm{u}_{\Delta t}^{(i)}\|_{L^2(\Omega)}^2
\end{equation*}
$\bm{\theta}$ is optimized using AdamW with learning rate scheduling~\cite{loshchilov2019decoupled}.

\FloatBarrier
\section{Implementation}
\label{sec:implementation}

\subsection{Dynamical Regimes and Parameter-Box Placement}
\label{sec:bifurcation}

To verify that the surrogate is trained on meaningful dynamics, we analyze the spatially-homogeneous reduction of \eqref{eq:fhn_u}--\eqref{eq:fhn_v}. One fixed point satisfies $u^\star - (u^\star)^3/3 = v^\star$ and $u^\star + a = b v^\star$, which yields the cubic $b (u^\star)^3 - 3(b-1) u^\star + 3 a = 0$. Each real root gives one fixed point, and its linear Jacobian
\begin{equation*}
J = \begin{pmatrix} 1 - (u^\star)^2 & -1 \\ 1/\tau & -b/\tau \end{pmatrix}
\end{equation*}
classifies the local flow. The regime is excitable when there is a single fixed point with $\mathrm{tr}\,J < 0$ and $\det J > 0$ (a stable rest state), and oscillatory when there is a single fixed point with $\mathrm{tr}\,J > 0$ and $\det J > 0$, an unstable focus surrounded by a limit cycle. It is bistable when there are three real fixed points.

\begin{figure}[!ht]
\centering
\includegraphics[width=\columnwidth]{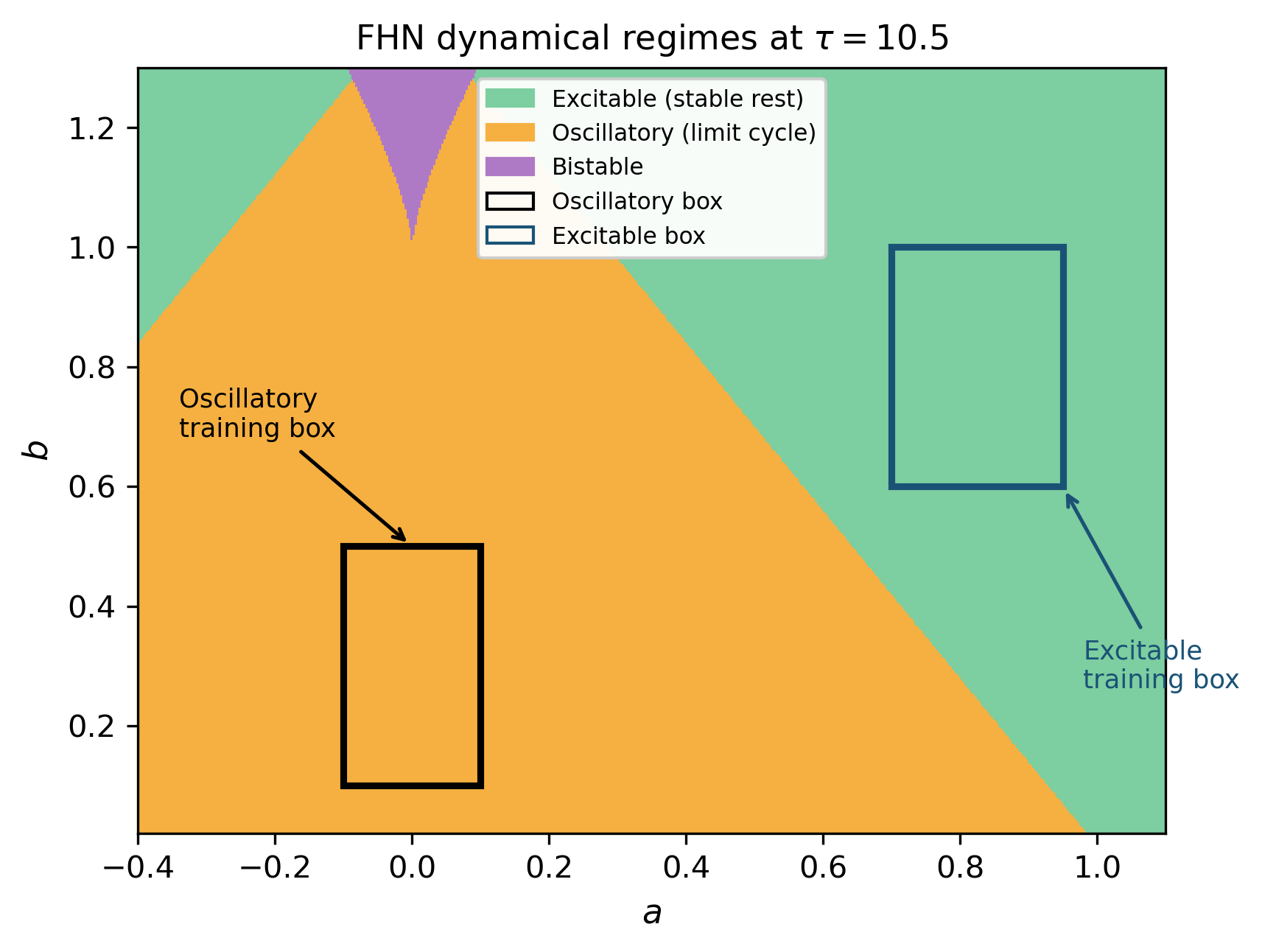}
\caption{Dynamical-regime map of the spatially-homogeneous FHN ODE at the median training $\tau = 10.5$. Green: excitable (stable rest state). Orange: oscillatory (limit cycle, tonic firing). Purple: bistable. The black rectangle is the oscillatory training box and the blue rectangle is the excitable training box. It is visually very clear, and supported by a sweep across the full $\tau$ range, that both boxes lie entirely within their respective regimes.}
\label{fig:bifurcation}
\end{figure}

\Cref{fig:bifurcation} maps these regimes on the $(a, b)$ plane at the median training $\tau$ and overlays the two parameter boxes this paper models. The first surrogate's box (\Cref{sec:data_generation}) clearly sits in the oscillatory region: a finer sweep over $\tau \in [1, 20]$ confirms $100\%$ of the sampled cube is oscillatory, so the surrogate is trained specifically on limit-cycle neurons firing tonically rather than a dormant rest state. These same criteria place the box of the excitable surrogate (\Cref{sec:excitable_results}, $|u^\star| > 1$) entirely in the stable-rest region. Thus, a single analysis defines both dynamical regimes this paper models.

\subsection{Data Generation}
\label{sec:data_generation}

We generate training and validation datasets by solving the FHN system \eqref{eq:fhn_u}-\eqref{eq:fhn_v} with a semi-implicit finite-difference method on a uniform periodic grid. We use second-order central differences for the Laplacian and treat the stiff diffusion term implicitly while the reaction term is explicit. Because the implicit operator $\mathbf{I} - \Delta t\, D\mathbf{L}$ is constant for a fixed parameter set, we precompute its LU factorization once per trajectory and reuse it at every step. 

\subsubsection{Initial Conditions}
\label{sec:initial_conditions}

We employ Gaussian Random Fields (GRF) as the standard initial condition type. In 1D, we compute Fourier coefficients according to:
\begin{align*}
    \hat{u}_k &= \mathcal{N}(0,1) \cdot (1 + |k|^2)^{-\alpha/2} \\
    \hat{v}_k &= \mathcal{N}(0,1) \cdot (1 + |k|^2)^{-\alpha/2} \cdot 0.5
\end{align*}
where $k$ is the wavenumber, $\mathcal{N}(0,1)$ denotes a standard normal random variable, and $\alpha = 2.0$ controls the spectral decay rate. The spatial fields are obtained via inverse Fourier transform, and we normalize each field. We manually add a 0.5 factor to the $v$ field to ensure the inhibitor variable has a smaller initial amplitude than the activator; this is supported by biophysical considerations~\cite{Izhikevich2007}.

\subsubsection{Parameter Sampling}
\label{sec:parameter_sampling}

All five parameters are sampled uniformly from physiologically motivated ranges~\cite{FitzHugh1961,Keener2009,Izhikevich2007}: $D_u \sim \mathcal{U}(0.01, 0.1)$ and $D_v \sim \mathcal{U}(0.005, 0.05)$ (voltage diffusing $\sim2\times$ faster than recovery, which is typical~\cite{Rinzel1998,Ermentrout2010}), $a \sim \mathcal{U}(-0.1, 0.1)$, $b \sim \mathcal{U}(0.1, 0.5)$, and $\tau \sim \mathcal{U}(1.0, 20.0)$ spans fast recovery to stiff relaxation oscillations. As established in \Cref{sec:bifurcation}, the entire sampled cube lies in the oscillatory regime of tonically firing neurons. The diffusion bounds span an order of magnitude, from near-pointwise dynamics at the lower bounds to the strong coupling that supports traveling waves at the upper bounds, with the ratio $D_u/D_v \approx 2$ that is common in FHN~\cite{Rinzel1998,Ermentrout2010}. Because $a$ is near zero, the fixed point is near the cubic nullcline's inflection where the Jacobian trace will produce relaxation oscillations. The recovery strength $b \in [0.1, 0.5]$ stays below the $b \approx 1$ that would restabilize a rest state, and $\tau \in [1, 20]$ spans near-equal to stiff fast/slow time scales~\cite{Izhikevich2007,Terman1992}.

\subsubsection{Dataset Specifications}
\label{sec:dataset_specs}

The primary dataset has $N = 8000$ trajectories on a periodic domain $\Omega = [0, 1]$ with $n_x = 256$ points, each evolved to $T = 1.0$ at $\Delta t = 0.01$ with $n_{\text{save}} = 50$ saved snapshots ($\alpha = 2.0$ GRF initial conditions). This yields 50 single-step pairs $(u_t, v_t) \rightarrow (u_{t+\Delta t}, v_{t+\Delta t})$ per trajectory under an 80-20 trajectory-level split (6{,}400 train / 1{,}600 validation) used for training and model selection. All final oscillatory-regime metrics in \Cref{sec:results} are instead reported on a separately generated, identically distributed held-out test set of $1{,}600$ trajectories, disjoint from both the training and validation data and unseen during training and hyperparameter selection. The excitable-regime experiments (\Cref{sec:excitable_results}) use a separately generated dataset on a longer domain $L = 8$, with $N = 1{,}250$ trajectories ($1{,}000$ train / $250$ validation), $T = 40$, $\Delta t = 0.02$, and $n_{\text{save}} = 100$, and localized super-threshold Gaussian-bump perturbations of the rest state as initial conditions rather than $\alpha = 2.0$ GRF fields.

\subsection{Model Architecture}
\label{sec:model_architecture}

We implement the FNO architecture as follows: the network retains the lowest $k_{\max} = 16$ Fourier modes in the convolution, uses a hidden dimension of $w = 64$, and consists of $L = 6$ Fourier layers. The input and output dimensions are both $d_{\text{in}} = d_{\text{out}} = 2$ to accommodate the coupled $(u, v)$ fields. We employ the GELU activation function for its smooth gradient properties, even though the computational cost is higher than the traditional ReLU activation function.

\subsubsection{Lifting and Projection Networks}
\label{sec:lifting_projection}

The lifting network $\mathcal{P}$ elevates the two-channel input to the $w$-dimensional hidden space through the following architecture:
\begin{equation*}
    \mathcal{P}(\bm{u}_0) = \text{Conv1D}_{64}\left(\text{GELU}\left(\text{Conv1D}_{128}(\bm{u}_0)\right)\right)
\end{equation*}
where $\text{Conv1D}_c$ denotes a kernel-size-1 1D convolution with $c$ output channels: the input is expanded to 128 channels, passed through GELU, then projected to the working width of 64.

Similarly, the projection network $\mathcal{Q}$ mirrors this structure to map from the hidden representation back to the physical $(u, v)$ space:
\begin{equation*}
    \mathcal{Q}(\bm{z}_L) = \text{Conv1D}_{2}\left(\text{GELU}\left(\text{Conv1D}_{64}(\bm{z}_L)\right)\right).
\end{equation*}
To preserve input features during the forward pass, we apply a global residual connection to the final output:
\begin{equation*}
    \bm{u}_{\text{pred}} = \mathcal{Q}(\bm{z}_L) + \beta_{\text{global}} \bm{u}_0
\end{equation*}
where $\beta_{\text{global}}$ is a learnable scalar initialized to 0.1. It allows the network to learn small perturbations to the input state rather than reconstructing the entire output from scratch.

\subsubsection{Fourier Layer Design}
\label{sec:fourier_layer_design}

Each of the $L = 6$ Fourier layers computes the real FFT $\tilde{\bm{z}} = \text{rfft}(\bm{z}_{\ell-1})$, retains only the first $k_{\max} = 16$ modes (we zero the rest which acts as an implicit low-pass filter). At the same time a pointwise $1\times1$ convolution captures local features, and the two pathways are summed and passed through GELU: $\bm{z}_{\text{pre}} = \text{GELU}(\bm{z}_{\text{spectral}} + \text{Conv1D}_{64}(\bm{z}_{\ell-1}))$.

\begin{figure}[!ht]
\centering
\includegraphics[width=\columnwidth]{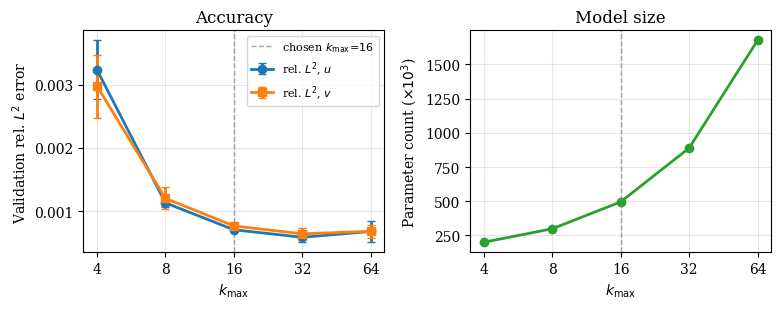}
\caption{Ablation of the Fourier-mode truncation $k_{\max}$ on the 8000-sample dataset (3 seeds per configuration). Left: validation rel-$L^2$ on $u$ and $v$. Right: parameter count. $k_{\max} = 16$ lies in the accuracy plateau while keeping model size modest.}
\label{fig:kmax_ablation}
\end{figure}

\paragraph{Choosing $k_{\max}{=}16$.} The diffusive terms $D_u \nabla^2 u$ and $D_v \nabla^2 v$ damp Fourier mode $k$ at a rate proportional to $k^2$, so even sharp FHN spikes concentrate their energy in the lowest few wavenumbers. An ablation over $k_{\max} \in \{4, 8, 16, 32, 64\}$ on the full 8000-trajectory dataset (\Cref{fig:kmax_ablation}) shows validation rel-$L^2$ improving $\sim 4{\times}$ from $k_{\max}=4$ to $k_{\max}=16$ and then plateauing, with $k_{\max}=64$ statistically indistinguishable from $k_{\max}=16$ while using $3.4{\times}$ more parameters. We also note that although $k_{\max}=32$ has a slightly better error score, the $k_{\max}=16$ model is more stable during autoregressive rollouts, which is the ultimate use case for the surrogate. Thus, we choose $k_{\max} = 16$ as a sweet spot. We ablate only $k_{\max}$, the axis the diffusive spectrum makes most interpretable. The hidden width $w = 64$ and depth $L = 6$ were fixed to standard FNO values~\cite{li2021fourierneuraloperatorparametric} that already reach sub-$0.1\%$ error and were not tuned further, so a fuller width/depth sweep remains future work.

\subsubsection{Parameter Conditioning}
\label{sec:parameter_conditioning}

The lifting, Fourier, and projection blocks described above only take $(u_t, v_t)$ as input, and so they cannot distinguish between different FHN systems. To make it parametric, we condition each Fourier layer on the parameter vector $\bm{\lambda} = (D_u, D_v, a, b, \tau)$ via feature-wise linear modulation (FiLM)~\cite{film2018}.

We first encode $\bm{\lambda}$ into a width-$w$ feature vector with a three-layer MLP $E$ with LayerNorm (widths $5 \to 2w \to w \to w$):
\begin{equation*}
    \bm{h}_{\bm{\lambda}} = E(\bm{\lambda}) \in \mathbb{R}^{w}.
\end{equation*}
At each Fourier layer $\ell$, two independent linear heads produce a per-channel scale $\bm{\gamma}_\ell \in \mathbb{R}^{w}$ and shift $\bm{\beta}_\ell \in \mathbb{R}^{w}$ from $\bm{h}_{\bm{\lambda}}$. The modulation is applied to the Fourier layer's output:
\begin{equation*}
    \bm{z}_\ell(\bm{x}) \leftarrow \bm{\gamma}_\ell \odot \bm{z}_\ell(\bm{x}) + \bm{\beta}_\ell
\end{equation*}
where $\odot$ denotes channel-wise multiplication. The encoder $E$ and the per-layer heads are trained with the rest of the network. This turns the single shared operator $\mathcal{G}_{\bm{\theta}}$ into a $\bm{\lambda}$-indexed family $\{\mathcal{G}_{\bm{\theta}, \bm{\lambda}}\}$, with modulation acting the same at every spatial location.

\subsection{Training and Evaluation}
\label{sec:training_config}

Training pairs $(\bm{u}_t, \bm{u}_{t+\Delta t})$ are drawn from trajectories with $u$ and $v$ independently $z$-normalized using statistics from the training set, and each batch also provides the per-trajectory parameter vector $\bm{\lambda}$. We optimize the per-element MSE over both fields using AdamW~\cite{loshchilov2019decoupled} with the following parameters: lr $10^{-3}$, weight decay $10^{-4}$, batch 32, gradient clipping 1.0, for up to 1000 epochs on an A100 GPU. We assess accuracy with the relative $L^2$ error per field, MSE, and the autoregressive rollout error obtained by iterating the single-step operator for up to $n_{\text{steps}} = 50$ consecutive steps. The parameter vector $\bm{\lambda}$ is standardized to zero mean and unit variance before entering the FiLM encoder. Five-seed runs underlie \Cref{tab:single_step_metrics} and the $\pm$std values in \Cref{tab:baselines}, and all other tables use seed $42$.

\begin{table}[!ht]
\centering
\small
\setlength{\tabcolsep}{6pt}
\renewcommand{\arraystretch}{1.05}
\begin{tabular}{@{}l cc@{}}
\toprule
 & \textbf{Oscillatory} & \textbf{Excitable} \\
\midrule
\multicolumn{3}{@{}l}{\textit{Data generation}} \\
\quad Grid $n_x$                 & $256$           & $256$        \\
\quad Domain $L$                 & $1.0$           & $8.0$        \\
\quad Time step $\Delta t$       & $0.01$          & $0.02$       \\
\quad Horizon $T$\,/\,frames     & $1.0$\,/\,$50$  & $40$\,/\,$100$ \\
\quad Train\,/\,val trajectories & $6400$\,/\,$1600$ & $1000$\,/\,$250$ \\
\addlinespace[2pt]
\multicolumn{3}{@{}l}{\textit{Architecture}} \\
\quad Fourier modes $k_{\max}$       & \multicolumn{2}{c}{$16$} \\
\quad Hidden width $w$               & \multicolumn{2}{c}{$64$} \\
\quad Fourier layers $L$             & \multicolumn{2}{c}{$6$}  \\
\quad Lift\,/\,project               & \multicolumn{2}{c}{$2\!\to\!128\!\to\!64$\,/\,$64\!\to\!2$} \\
\quad FiLM encoder                   & \multicolumn{2}{c}{$5\!\to\!128\!\to\!64\!\to\!64$} \\
\quad Global skip $\beta_{\text{global}}$ init & \multicolumn{2}{c}{$0.1$} \\
\quad Parameters                     & \multicolumn{2}{c}{${\sim}0.50$\,M} \\
\bottomrule
\end{tabular}
\smallskip
\caption{Data-generation, architecture, and training hyperparameters.}
\label{tab:hyperparams}
\end{table}

\FloatBarrier
\section{Results}
\label{sec:results}

We evaluate our with five different metrics. We first establish single-step accuracy (is one operator application faithful?) and further analyze the errors on a per-parameter basis. We then examine a long-horizon autoregressive rollout (does error stay bounded when the operator is iterated in long-duration simulations, the actual use case?), as well as generalization outside the training parameter ranges. Finally, we evaluate the computational efficiency of the FNO surrogate compared to finite-difference solvers.

Unless stated otherwise, the results in this section concern the oscillatory (tonic-firing) regime; the comparably accurate results for the complementary excitable regime can be found in \Cref{sec:excitable_results}. Model and hyperparameter selection (e.g.\ the $k_{\max}$ ablation of \Cref{sec:fourier_layer_design}) used the validation split, whereas all errors reported here are computed on the disjoint held-out test set of $1{,}600$ trajectories described in \Cref{sec:dataset_specs}, whose parameter combinations and initial conditions were unseen during training and selection.

\subsection{Single-Step Prediction Accuracy}
\label{sec:single_step_results}

We begin by evaluating the FNO's ability to predict single time steps. Table~\ref{tab:single_step_metrics} summarizes the quantitative performance metrics averaged over the entire test set.

\begin{table}[!ht]
\centering
\begin{small}
\begin{sc}
\setlength{\tabcolsep}{4pt}
\begin{tabular}{lcc}
\toprule
Metric & Activator ($u$) & Inhibitor ($v$) \\
\midrule
Rel.\ $L^2$ & $6.66 \times 10^{-4}$ & $8.01 \times 10^{-4}$ \\
MSE & $2.98 \times 10^{-6}$ & $9.05 \times 10^{-6}$ \\
MAE & $4.09 \times 10^{-4}$ & $5.86 \times 10^{-4}$ \\
Max AE & $0.178$ & $0.309$ \\
\bottomrule
\end{tabular}
\end{sc}
\end{small}
\smallskip
\caption{Single-step prediction accuracy on the held-out test set ($N_{\text{test}} = 1600$ trajectories, $n_x = 256$ spatial points), in normalized space. Values are 5-seed means (per-trajectory mean for relative $L^2$, element-wise averages otherwise). Per-seed variability is reported in \Cref{tab:baselines}.}
\label{tab:single_step_metrics}
\end{table}

The parametric FNO predicts single time steps with sub-$0.1\%$ relative $L^2$ error on both fields: $6.66 \times 10^{-4}$ ($0.067\%$) for the activator $u$ and $8.01 \times 10^{-4}$ ($0.080\%$) for the inhibitor $v$. The inhibitor's marginally higher relative error is expected given its smaller amplitude, so that comparable absolute deviations then translate into larger relative values. Mean absolute errors ($4.1 \times 10^{-4}$ for $u$, $5.9 \times 10^{-4}$ for $v$) and mean squared errors ($3.0 \times 10^{-6}$, $9.0 \times 10^{-6}$) are physically negligible, and the largest pointwise deviations of max absolute error $0.18$ for $u$ and $0.31$ for $v$ are confined to sharp wavefronts, leaving the total metrics almost entirely unaffected.

\begin{figure}[!ht]
\centering
\includegraphics[width=\columnwidth]{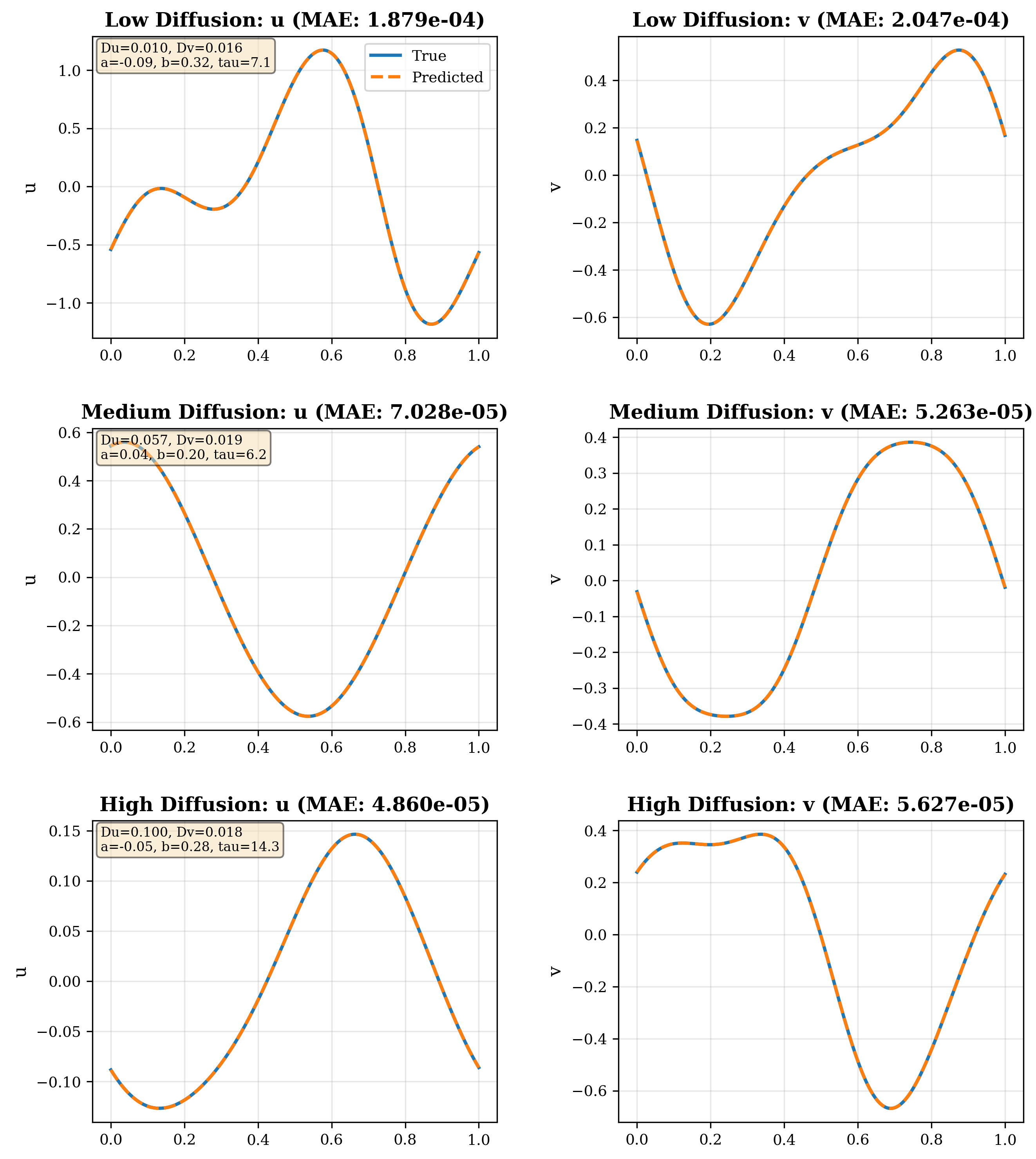}
\caption{Single-step prediction across three representative test regimes (rows: weakly, moderately, strongly diffusive). Left: ground truth vs predicted $u$. Center: ground truth vs predicted $v$. Right: absolute error in $u$ and $v$ (scaled for visibility). Predicted and true profiles are visually indistinguishable.}
\label{fig:single_step_comparison}
\end{figure}

Qualitative single-step predictions across weakly, moderately, and strongly diffusive regimes are visually indistinguishable from ground truth (\Cref{fig:single_step_comparison}). The per-regime single-step MAE ranges from $5 \times 10^{-5}$ to $2 \times 10^{-4}$, with the weakly diffusive case having the largest errors, consistent with the diffusion-driven failure mode of \Cref{sec:extrapolation_results}. It notes that weaker diffusion leaves sharper wavefronts that are truncated by the $k_{\max}{=}16$ spectral filter.

We additionally compare against two natural baselines. ``Baseline FNO'' uses the identical backbone (width 64, 16 modes, 6 layers) but with the parameter vector $\bm{\lambda}$ broadcast as five constant spatial input channels instead of being injected via FiLM. Because the two models share architecture, capacity, and training and differ only in how $\bm{\lambda}$ enters, this baseline isolates the effect of the FiLM conditioning itself rather than of added capacity. DeepONet~\cite{lu2021learning} is a operator-learning architecture that encodes the input field and the query coordinates through separate subnetworks. The DeepONet comparison mode uses a flattened $(u_t, v_t)$ concatenated with $\bm{\lambda}$. The results are shown in Table~\ref{tab:baselines}. DeepONet performs far worse here: its relative $L^2$ error is roughly $58\times$ higher on $u$ and $50\times$ higher on $v$. The channel-broadcast FNO is more competitive but still performs worse than the FiLM-conditioned model by roughly $30\%$ on both fields, while still containing $26\%$ more parameters ($0.63$M vs $0.50$M). FiLM conditioning therefore improves accuracy and reduces model size relative to broadcasting $\bm{\lambda}$ as constant channels, so the gain comes from how parameters are added rather than from added capacity.

\begin{table*}[!ht]
\begin{center}
\begin{scriptsize}
\begin{sc}
\setlength{\tabcolsep}{5pt}
\begin{tabular}{lccccc}
\toprule
 & \multicolumn{2}{c}{$\epsilon_{\text{rel}}$ ($\times 10^{-3}$)} & \multicolumn{2}{c}{MAE ($\times 10^{-3}$)} & \\
\cmidrule(lr){2-3} \cmidrule(lr){4-5}
Method & $u$ & $v$ & $u$ & $v$ & Params \\
\midrule
DeepONet
  & $38.3 \pm 2.0$ & $40.3 \pm 3.4$
  & $25.4 \pm 2.1$ & $28.8 \pm 2.5$
  & $0.40$M \\
Base FNO
  & $0.859 \pm 0.046$ & $1.038 \pm 0.059$
  & $0.798 \pm 0.019$ & $0.898 \pm 0.033$
  & $0.63$M \\
Param. FNO
  & $\bm{0.666 \pm 0.080}$ & $\bm{0.801 \pm 0.140}$
  & $\bm{0.409 \pm 0.022}$ & $\bm{0.586 \pm 0.098}$
  & $0.50$M \\
\bottomrule
\end{tabular}
\end{sc}
\end{scriptsize}
\end{center}
\smallskip
\caption{Baseline comparison on the test set ($N_{\text{test}} = 1600$
trajectories, $n_x = 256$). All error values are mean $\pm$ standard deviation
across $n_{\text{seeds}} = 5$ independent training runs, reported in units of
$10^{-3}$. }
\label{tab:baselines}
\end{table*}

\subsection{Parameter Error Correlation}
\label{sec:parameter_generalization}

It's important to note that a single averaged error can hide some behavior: a surrogate may look accurate overall while failing along one axis of the parameter space. Because the coordinates of $\bm{\lambda}$ correspond to distinct ODE bifurcations, the axes where error concentrates carry physical meaning and indicate where to add data augmentation. Here, we relate the error to each axis of $\bm{\lambda}$ individually so we can determine whether any single parameter drives accuracy. We note that the Pearson $\rho$ used here captures only linear/monotonic dependence, so a purely nonlinear sensitivity could go undetected. Table~\ref{tab:parameter_correlation} reports Pearson correlations between each FHN parameter and the relative $L^2$ error for both $u$ and $v$.

\begin{table}[!ht]
\begin{center}
\begin{small}
\begin{sc}
\begin{tabular}{lccccc}
\toprule
 & $D_u$ & $D_v$ & $a$ & $b$ & $\tau$ \\
\midrule
$u$ error & $+0.143$ & $-0.047$ & $-0.036$ & $+0.033$ & $-0.052$ \\
$v$ error & $-0.166$ & $+0.114$ & $+0.025$ & $-0.027$ & $-0.357$ \\
\bottomrule
\end{tabular}
\end{sc}
\end{small}
\end{center}
\smallskip
\caption{Pearson correlation between each FHN parameter and the per-trajectory single-step relative $L^2$ error on the test set. All $|\rho| < 0.36$, indicating relatively uniform generalization across the five-dimensional parameter space.}
\label{tab:parameter_correlation}
\end{table}

The largest correlation magnitude anywhere in \Cref{tab:parameter_correlation} is $|\rho| = 0.357$ (the $\tau$--$v$ term), which indicates relatively uniform accuracy. Contrary to expectation that the faster activator would be more sensitive, $u$ is nearly parameter-agnostic: its strongest correlation is a weak $\rho = +0.143$ with $D_u$, and every other coefficient is under $|\rho| \le 0.06$. The variation that exists is concentrated in the inhibitor $v$, whose dominant axis is the time-scale $\tau$, with $\rho = -0.357$. This is physically sensible, since $\tau$ sets the relaxation rate of the slow recovery variable. The negative sign means $v$ is predicted slightly more accurately at larger $\tau$, where its dynamics are smoother. The reaction parameters $a$ and $b$ are basically uncorrelated with the error on both fields, with $|\rho| \le 0.05$.

Error thus concentrates along the time-scale axis rather than the reaction axes, but even there the dependence is moderate. This relatively uniform structure indicates that random parameter sampling sufficed to cover the space, and the within-range trend that higher $\tau$ is easier is consistent with the time-scale extrapolation shown next.

\subsection{Long-Duration Autoregressive Rollout}
\label{sec:autoregressive_results}

Many applications require predictions over extended time horizons. We evaluate the FNO's autoregressive rollout capability by iteratively applying the single-step operator for $n_{\text{steps}} = 50$ time steps, corresponding to a total evolution time of $T = 0.5$ time units ($50\times$ the training time step $\Delta t = 0.01$).

Most substantially, the operator does not suffer from runaway error accumulation: even after $50$ autoregressive steps the relative $L^2$ error is only $\sim$$0.5\%$ for the activator $u$ and $\sim$$0.7\%$ for the inhibitor $v$. The activator error grows roughly linearly, whereas the inhibitor error drops over the first few steps before increasing sub-linearly. This early dip reflects the inhibitor's slow time scale, as the single-step operator slightly over-sharpens $v$, but $v$'s own slow, diffusive relaxation damps that transient over the next few steps before the regular error accumulation takes over. Because the activator error grows faster, the two curves converge by step $50$ and $u$ would likely surpass $v$ over a longer time horizon.

\begin{figure}[!ht]
\centering
\includegraphics[width=\columnwidth]{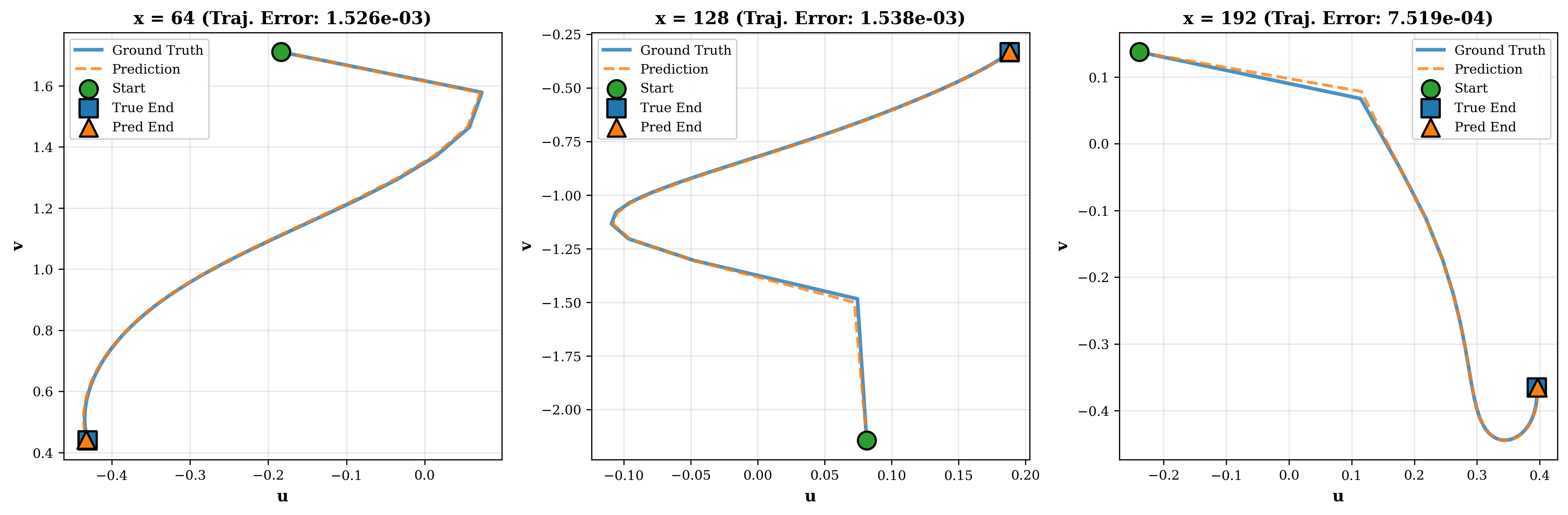}
\caption{Phase-space portraits at three spatial locations ($v$ vs $u$, $50$ time steps). Blue: ground truth, Orange dashed: FNO. Green circle: initial condition. Square and triangle: ground-truth and predicted endpoints, respectively. The operator traces the same limit-cycle geometry as the solver, with only a gradual end-of-rollout phase drift.}
\label{fig:phase_portraits}
\end{figure}

Note that over longer time horizons the FNO error could grow to concerning levels. Future work could explore training strategies such as fixed-horizon rollouts to improve long-term stability. A phase-space analysis (\Cref{fig:phase_portraits}) confirms that the FNO preserves the limit-cycle geometry at fixed spatial locations, with the only departure being an extremely slow phase drift consistent with the autoregressive error accumulation.

\subsection{Generalization to Unseen Parameters}
\label{sec:extrapolation_results}
We evaluate the FNO on parameters outside the training ranges to determine its extrapolation capabilities. Table~\ref{tab:extrapolation} reports four scenarios.

\begin{table}[!htb]
\centering
\begin{sc}
\fontsize{8}{9.6}\selectfont
\setlength{\tabcolsep}{4pt}
\begin{tabular}{lcc}
\toprule
Parameter Regime & $\epsilon_{\text{rel}}(u)$ & $\epsilon_{\text{rel}}(v)$ \\
\midrule
Within range (test) & $0.0006655$ & $0.0008012$ \\
Low diffusion ($D_u{=}0.005$, $D_v{=}0.002$) & $0.007085$ & $0.008089$ \\
High diffusion ($D_u{=}0.15$, $D_v{=}0.08$) & $0.03401$ & $0.01635$ \\
High time-scale separation ($\tau{=}30$) & $0.0005277$ & $0.001066$ \\
Low recovery ($b{=}0.05$) & $0.0003559$ & $0.0003577$ \\
\bottomrule
\end{tabular}
\end{sc}
\smallskip
\caption{Generalization performance outside the training ranges. Relative $L^2$ error, mean over 100 test trajectories per scenario.}
\label{tab:extrapolation}
\end{table}

Diffusion magnitude is the primary failure mode. When both diffusion coefficients are $50$--$60\%$ above the training range, the activator error increases to $3.4\%$ (roughly $51\times$ the within-range value) and the inhibitor to $1.6\%$ ($\sim$$20\times$), while lowering them below the training range gives only a $\sim$$10\times$ increase. In contrast, the model extrapolates well along the reaction and time-scale axes: while at $\tau{=}30$, the activator error stays below the within-range value and the inhibitor rises only by $\sim$$33\%$, and the low-recovery case ($b{=}0.05$) is more accurate than within-range on both fields. This contradicts the idea that the stiff time-scale separation regime is the difficult one. Instead, the sharp spatial gradients, which shift energy into the high wavenumbers that become truncated by the spectral layers, are what degrade accuracy. This suggests that future work could improve extrapolation by augmenting the training data with sharper profiles.

\subsection{Computational Efficiency}
\label{sec:efficiency_results}

Table~\ref{tab:efficiency_comparison} compares the FNO against the semi-implicit FD solver on one 50-step trajectory rollout. The FNO delivers $111\times$ speedup at batch 1 and $920\times$ at batch 32, turning a 2-second simulation into 2 ms. There is, however, a tradeoff: $\sim$150 MB for the FNO vs 8 MB for the FD solver's sparse matrices. However, we think this is a reasonable price for the speedup.

\begin{table}[!ht]
\centering
\begin{small}
\setlength{\tabcolsep}{4pt}
\begin{tabular}{lccc}
\toprule
Method & Time/step & Total (50 steps) & Memory \\
\midrule
FD (batch 32) & $42.3$ ms & $2.12$ s & $8.4$ MB \\
FNO (batch 1) & $0.38$ ms & $0.019$ s & $147$ MB \\
FNO (batch 32) & $0.045$ ms & $0.0023$ s & $152$ MB \\
\midrule
Speedup & $111\times$ / $940\times$ & $111\times$ / $920\times$ & --- \\
\bottomrule
\end{tabular}
\end{small}
\smallskip
\caption{Computational efficiency comparison between FNO and finite-difference (FD) solver for a single trajectory rollout ($n_x = 256$, $n_{\text{steps}} = 50$, NVIDIA A100).}
\label{tab:efficiency_comparison}
\end{table}

\subsection{Excitable Regime: Action-Potential Propagation}
\label{sec:excitable_results}

The experiments above target the oscillatory regime. A complementary and historically defining regime is the excitable one: a single stable rest state from which a sufficiently strong localized stimulus launches a traveling action potential. It exhibits two phenomena absent from the limit-cycle regime, an all-or-none firing threshold and a diffusion-controlled conduction velocity, and we test whether the same FiLM-conditioned operator learns them.

We train a second, architecturally identical surrogate (\Cref{tab:hyperparams}) on a parameter box that \Cref{sec:bifurcation} places entirely in the excitable region. A fixed point is excitable whenever $|u^\star| > 1$, and sampling $D_u \in [0.006, 0.030]$, $D_v \in [0.0005, 0.005]$, $a \in [0.70, 0.95]$, $b \in [0.60, 1.00]$, $\tau \in [8, 20]$ yields $|u^\star| \ge 1.10$ for all $5{\times}10^4$ Monte-Carlo evaluations. Pulses are launched from rest by a super-threshold Gaussian bump on a domain $L = 8$, which is long enough to keep the counter-propagating fronts from colliding with each other.

\begin{figure}[!htbp]
\centering
\includegraphics[width=\columnwidth]{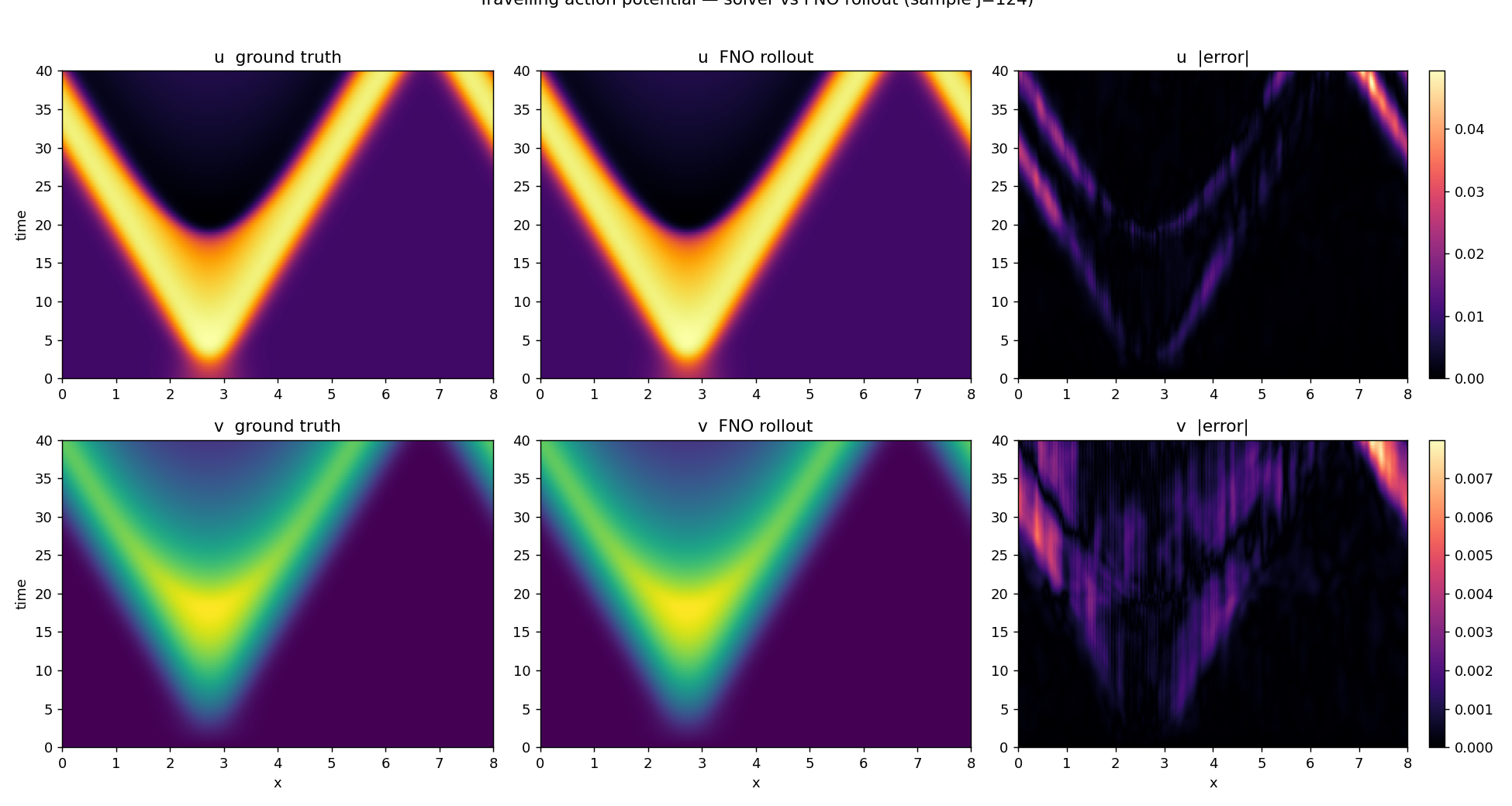}
\caption{Traveling action potential, solver vs FNO rollout. Rows: activator $u$ (top), recovery $v$ (bottom). Columns: ground truth, FNO $100$-step rollout, absolute error. The propagating wavefront is reproduced over the full $T = 40$ horizon, and error is confined to the moving front.}
\label{fig:excitable_rollout}
\end{figure}

Single-step accuracy matches the oscillatory model (\Cref{tab:excitable_metrics}): relative $L^2 \le 2 \times 10^{-4}$ on both fields. A $100$-step rollout over the full $T = 40$ horizon keeps relative $L^2$ below $0.9\%$, thus the model is able to serve as a surrogate for the entire traveling pulse rather than simply its onset (\Cref{fig:excitable_rollout}).

\begin{table}[!ht]
\centering
\begin{small}
\begin{sc}
\begin{tabular}{lcc}
\toprule
Metric & Activator ($u$) & Inhibitor ($v$) \\
\midrule
Relative $L^2$ & $1.97 \times 10^{-4}$ & $1.87 \times 10^{-4}$ \\
MSE & $1.5 \times 10^{-7}$ & $1.7 \times 10^{-8}$ \\
MAE & $1.8 \times 10^{-4}$ & $6.9 \times 10^{-5}$ \\
Max AE & $0.026$ & $0.0039$ \\
\bottomrule
\end{tabular}
\end{sc}
\end{small}
\smallskip
\caption{Single-step prediction accuracy in the excitable regime, $N = 250$ trajectories, in normalized units.}
\label{tab:excitable_metrics}
\end{table}

\begin{figure}[!ht]
\centering
\begin{subfigure}{0.49\columnwidth}
    \centering
    \includegraphics[width=\linewidth]{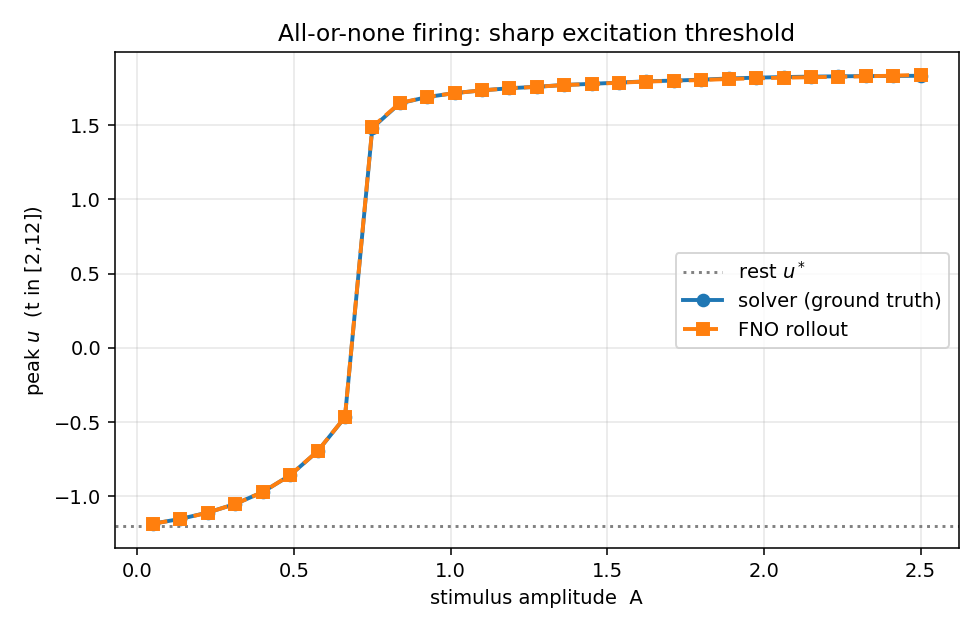}
    \caption{All-or-none threshold.}
    \label{fig:excitable_threshold}
\end{subfigure}
\hfill
\begin{subfigure}{0.49\columnwidth}
    \centering
    \includegraphics[width=\linewidth]{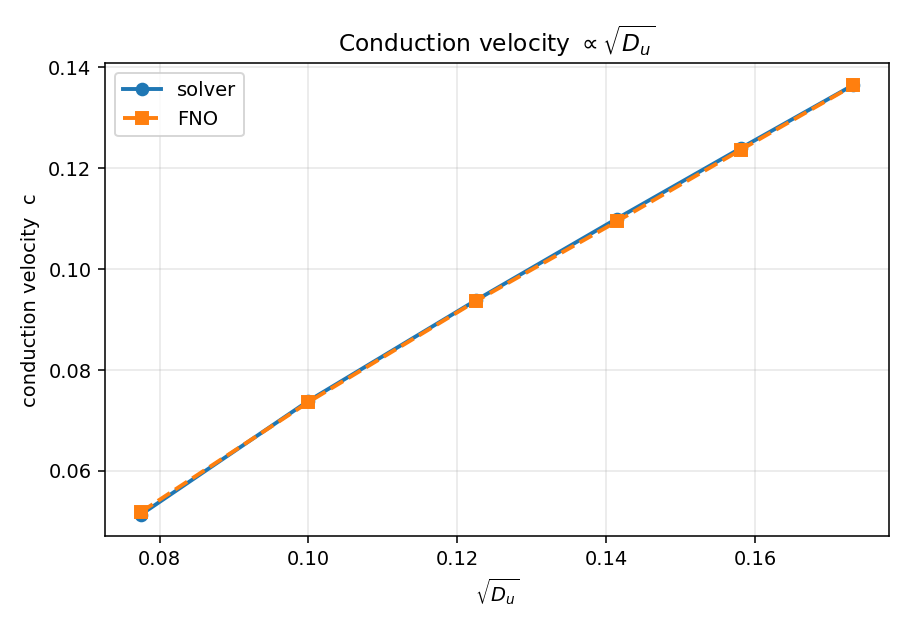}
    \caption{Conduction velocity.}
    \label{fig:excitable_cv}
\end{subfigure}
\caption{Quantitative excitable-media signatures, solver vs.\ FNO rollout. \textbf{(a)} Peak post-stimulus voltage vs stimulus amplitude $A$: a sharp threshold ($A^\star \approx 0.7$) separates sub-threshold decay to rest ($u^\star \approx -1.2$) from full action-potential firing ($u \approx +1.8$), and both methods are identical on both branches and at the threshold. \textbf{(b)} Traveling-pulse speed scales linearly with $\sqrt{D_u}$ as demonstrated by excitable-media theory, and the FNO matches the solver perfectly.}
\label{fig:excitable_signatures}
\end{figure}

Both excitable-media signatures match the solver (\Cref{fig:excitable_signatures}). The all-or-none threshold is visible as a sharp saddle-node transition at $A^\star \approx 0.7$ \Cref{fig:excitable_threshold}, and the conduction velocity follows the predicted $c \propto \sqrt{D_u}$ law~\cite{Keener2009} across the full diffusion sweep (\Cref{fig:excitable_cv}). Over the full $100$-step horizon the operator reproduces both fields of the propagating pulse, with error confined to a thin band at the moving wavefront, so it clearly tracks the live pulse rather than only its onset (\Cref{fig:excitable_rollout}).

\subsection*{Limitations and Future Work}
\label{sec:outlook}
Several limitations bound the present study and set up its natural extensions. First, all experiments are one-dimensional; we leave 2D/3D geometries for future work. Second, extrapolation degrades primarily along the diffusion axis (\Cref{sec:extrapolation_results}), where sharp wavefronts push energy past the spectral truncation. Third, the parameter-sensitivity screen of \Cref{sec:parameter_generalization} is linear and could miss nonlinear dependencies.

Most importantly, the differentiable surrogate is meant to enable downstream inverse and control tasks that we motivate. A concrete target is signal-based conduction blocking: finding a stimulus $I_{\text{ext}}(\bm{x}, t)$ that halts action-potential propagation, the basis of kHz nerve block~\cite{Kilgore2014}. We establish two prerequisites for this. (i)~A differentiable, accurate forward surrogate: because the FNO is differentiable in $\bm{\lambda}$ and its inputs, such a stimulus becomes a gradient-descent target, with the search direction obtained in a single backward pass through the rollout, whereas the numerical solver requires a separate integration for every candidate. (ii)~Parameter robustness: since the FiLM conditioning already spans the parameter family, a candidate signal can be stress-tested for robustness across $\bm{\lambda}$. Demonstrating an actual recovered-parameter or optimized-stimulus result is the natural next step, moving the surrogate toward equation-free, system-level neuromodulation design~\cite{fabiani2025enabling}.

\FloatBarrier
\section*{Ethical Statement}

Training neural operators is computationally intensive and carries a notable environmental footprint. While this FNO model is far smaller than state-of-the-art systems, the concern remains: the cumulative energy used in hyperparameter tuning, model training, and extensive simulations can be non-negligible. We plan to mitigate the environmental impact by limiting the number of training runs and using mixed-precision and tuned batch sizes to maximize hardware utilization \cite{patterson2022carbonfootprintmachinelearning}.

In addition, over-reliance on AI-generated approximations in high-stakes domains can be dangerous. If FNO modeling of the FHN system is used as a building block for general research (e.g., building more refined neuronal models), and researchers place blind trust in the predictions, it may lead to incorrect features or research based on false data. Overdependence on AI tools can lower human expertise and become an issue when the AI fails or is used outside its scope, especially in the medical field \cite{korkmaz2024artificial}. Thus, the FNO models should augment decision making rather than replace it.

Finally, the bias in modeling from training data or underlying design is another ethical concern. In the context of FNOs for the FitzHugh-Nagumo model, a form of bias could arise if the training dataset of simulated scenarios is not sufficiently representative of all relevant conditions (for example, if all training simulations use a narrow range of model parameters or initial conditions). It is known that when training data are unrepresentative or incomplete, the learned model will yield biased outputs that systematically err on those underrepresented conditions \cite{ferrara2023fairness}. If the model is applied in a biomedical context, it should be cross-checked to make sure it does not inadvertently perpetuate any biases that could lead to health disparities, i.e., differing accuracy on data from different patient groups.

\pagebreak

\bibliography{references}

@article{FitzHugh1961,
  author    = {FitzHugh, R.},
  title     = {Impulses and physiological states in theoretical models of nerve membrane},
  journal   = {Biophysical Journal},
  volume    = {1},
  number    = {6},
  pages     = {445--466},
  year      = {1961}
}

@article{Nagumo1962,
  author    = {Nagumo, J. and Arimoto, S. and Yoshizawa, S.},
  title     = {An active pulse transmission line simulating nerve axon},
  journal   = {Proceedings of the IRE},
  volume    = {50},
  number    = {10},
  pages     = {2061--2070},
  year      = {1962}
}

@book{Izhikevich2007,
  author    = {Izhikevich, E. M.},
  title     = {Dynamical Systems in Neuroscience: The Geometry of Excitability and Bursting},
  publisher = {MIT Press},
  year      = {2007}
}

@book{Keener2009,
  author    = {Keener, J. and Sneyd, J.},
  title     = {Mathematical Physiology},
  publisher = {Springer},
  year      = {2009}
}

@incollection{Rinzel1998,
  author       = {Rinzel, J. and Ermentrout, G. B.},
  title        = {Analysis of neural excitability and oscillations},
  booktitle    = {Methods in Neuronal Modeling},
  publisher    = {MIT Press},
  pages        = {251--291},
  year         = {1998}
}

@book{Ermentrout2010,
  author    = {Ermentrout, G. B. and Terman, D. H.},
  title     = {Mathematical Foundations of Neuroscience},
  publisher = {Springer},
  year      = {2010}
}

@article{Terman1992,
  author    = {Terman, D.},
  title     = {The transition from bursting to continuous spiking in excitable membrane models},
  journal   = {Journal of Nonlinear Science},
  volume    = {2},
  number    = {2},
  pages     = {135--182},
  year      = {1992},
  doi       = {10.1007/BF02429854}
}

@book{Quarteroni2000,
  author    = {Quarteroni, A. and Sacco, R. and Saleri, F.},
  title     = {Numerical Mathematics},
  publisher = {Springer},
  year      = {2000}
}

@book{LeVeque2007,
  author    = {LeVeque, R. J.},
  title     = {Finite Difference Methods for Ordinary and Partial Differential Equations: Steady-State and Time-Dependent Problems},
  publisher = {SIAM},
  year      = {2007}
}

@ARTICLE{2020PhRvR...2b3068B,
       author = {{Burns}, Keaton J. and {Vasil}, Geoffrey M. and {Oishi}, Jeffrey S. and {Lecoanet}, Daniel and {Brown}, Benjamin P.},
        title = "{Dedalus: A flexible framework for numerical simulations with spectral methods}",
      journal = {Physical Review Research},
         year = 2020,
        month = apr,
       volume = {2},
       number = {2},
          eid = {023068},
        pages = {023068},
          doi = {10.1103/PhysRevResearch.2.023068},
archivePrefix = {arXiv},
       eprint = {1905.10388},
 primaryClass = {astro-ph.IM},
       adsurl = {https://ui.adsabs.harvard.edu/abs/2020PhRvR...2b3068B}
}

@article{kovachki2021neural,
  author = {Nikola B. Kovachki and Zongyi Li and Burigede Liu and Kamyar Azizzadenesheli and Kaushik Bhattacharya and Andrew M. Stuart and Anima Anandkumar},
  title = {Neural Operator: Learning Maps Between Function Spaces},
  journal = {CoRR},
  volume = {abs/2108.08481},
  year = {2021},
}

@misc{li2021fourierneuraloperatorparametric,
      title={Fourier Neural Operator for Parametric Partial Differential Equations},
      author={Zongyi Li and Nikola Kovachki and Kamyar Azizzadenesheli and Burigede Liu and Kaushik Bhattacharya and Andrew Stuart and Anima Anandkumar},
      year={2021},
      eprint={2010.08895},
      archivePrefix={arXiv},
      primaryClass={cs.LG},
      url={https://arxiv.org/abs/2010.08895},
}

@article{lu2021learning,
  title={Learning nonlinear operators via {DeepONet} based on the universal approximation theorem of operators},
  author={Lu, Lu and Jin, Pengzhan and Pang, Guofei and Zhang, Zhongqiang and Karniadakis, George Em},
  journal={Nature Machine Intelligence},
  volume={3},
  number={3},
  pages={218--229},
  year={2021},
  publisher={Nature Publishing Group}
}

@article{grady2023modelparallel,
  title={Model-parallel {Fourier} neural operators as learned surrogates for large-scale parametric {PDEs}},
  author={Grady, Thomas J. and Khan, Rishi and Louboutin, Mathias and Yin, Ziyi and Witte, Philipp A. and Chandra, Ranveer and Hewett, Russell J. and Herrmann, Felix J.},
  journal={Computers \& Geosciences},
  volume={178},
  pages={105402},
  year={2023},
  doi={10.1016/j.cageo.2023.105402}
}

@article{ghadjari2026fno,
  title={A {Fourier} neural operator surrogate model for nonlinear electrical resistivity tomography},
  author={Ghadjari, H. and Shahsavari, P. and Dettmer, J. and Azizzadenesheli, K. and Gilbert, H.},
  journal={Geophysical Journal International},
  volume={244},
  number={3},
  year={2026},
  doi={10.1093/gji/ggag022}
}

@article{kang2024adjoint,
  title={Adjoint method-based {Fourier} neural operator surrogate solver for wavefront shaping in tunable metasurfaces},
  author={Kang, Chanik and Seo, Joonhyuk and Jang, Ikbeom and Chung, Haejun},
  journal={iScience},
  volume={28},
  number={1},
  year={2025},
  doi={10.1016/j.isci.2024.111545},
  url={https://www.cell.com/iscience/fulltext/S2589-0042(24)02772-X}
}

@article{hao2024fourier,
  title={Fourier Neural Operator Networks for Solving Reaction--Diffusion Equations},
  author={Hao, Yaobin and Song, Fangying},
  journal={Fluids},
  volume={9},
  number={11},
  pages={258},
  year={2024},
  publisher={MDPI}
}

@article{fabiani2025enabling,
  title={Enabling Local Neural Operators to perform Equation-Free System-Level Analysis},
  author={Fabiani, Gianluca and Vandecasteele, Hannes and Goswami, Somdatta and Siettos, Constantinos and Kevrekidis, Ioannis G},
  journal={arXiv preprint arXiv:2505.02308},
  year={2025}
}

@inproceedings{loshchilov2019decoupled,
  title={Decoupled Weight Decay Regularization},
  author={Ilya Loshchilov and Frank Hutter},
  booktitle={International Conference on Learning Representations (ICLR)},
  year={2019},
  url={https://openreview.net/forum?id=Bkg6RiCqY7}
}

@inproceedings{film2018,
  title={{FiLM}: Visual Reasoning with a General Conditioning Layer},
  author={Ethan Perez and Florian Strub and Harm de Vries and Vincent Dumoulin and Aaron Courville},
  booktitle={AAAI Conference on Artificial Intelligence},
  year={2018},
  url={https://arxiv.org/abs/1709.07871}
}

@misc{patterson2022carbonfootprintmachinelearning,
      title={The Carbon Footprint of Machine Learning Training Will Plateau, Then Shrink},
      author={David Patterson and Joseph Gonzalez and Urs Hölzle and Quoc Le and Chen Liang and Lluis-Miquel Munguia and Daniel Rothchild and David So and Maud Texier and Jeff Dean},
      year={2022},
      eprint={2204.05149},
      archivePrefix={arXiv},
      primaryClass={cs.LG},
      url={https://arxiv.org/abs/2204.05149},
}

@article{korkmaz2024artificial,
  title={Artificial intelligence in healthcare: a revolutionary ally or an ethical dilemma?},
  author={Korkmaz, Sel{\c{c}}uk},
  journal={Balkan Medical Journal},
  volume={41},
  number={2},
  pages={87--88},
  year={2024}
}

@article{Kilgore2014,
  author    = {Kilgore, K. L. and Bhadra, N.},
  title     = {Reversible nerve conduction block using kilohertz frequency alternating current},
  journal   = {Neuromodulation: Technology at the Neural Interface},
  volume    = {17},
  number    = {3},
  pages     = {242--254},
  year      = {2014}
}

@article{ferrara2023fairness,
  title={Fairness and bias in artificial intelligence: A brief survey of sources, impacts, and mitigation strategies},
  author={Ferrara, Emilio},
  journal={Sci},
  volume={6},
  number={1},
  pages={3},
  year={2023},
  publisher={MDPI}
}
\bibliographystyle{splncs04}

\end{document}